\pdfoutput=1

\documentclass[sigconf]{acmart}

\usepackage{amssymb,amsmath,amsfonts}
\usepackage{algorithm}
\usepackage[position=bottom,caption=false,font=footnotesize,labelfont=rm,textfont=rm]{subfig}
\usepackage{textcomp}
\usepackage{stfloats}
\usepackage{graphicx}
\usepackage{float}
\usepackage{multirow}
\usepackage{color}
\usepackage{hyperref}
\usepackage{bbding}
\usepackage{balance}

\AtBeginDocument{%
  }

\copyrightyear{2025}
\acmYear{2025}
\setcopyright{acmlicensed} 
\acmConference[SVC '25] {Proceedings of the 1st International Workshop \& Challenge on Subtle Visual Computing}{October 27--31, 2025}{Dublin, Ireland.}
\acmBooktitle{Proceedings of the 1st International Workshop \& Challenge on Subtle Visual Computing (SVC '25), October 27--31, 2025, Dublin, Ireland}
\acmISBN{979-8-4007-1837-3/2025/10}
\acmDOI{10.1145/3728425.3759924}

\begin{document}

\title{Multi-source Multimodal Progressive Domain Adaption for Audio-Visual Deception Detection}


\author{Ronghao Lin}
\orcid{0000-0003-4530-4529}
\affiliation{%
  \institution{Sun Yat-Sen University}
  \city{Guangzhou}
  \country{China}}
\affiliation{%
  \institution{Nanyang Technological University}
  \country{Singapore}}
\email{linrh7@mail2.sysu.edu.cn}

\author{Sijie Mai}
\orcid{0000-0001-9763-375X}
\affiliation{%
  \institution{South China Normal University}
  \city{Guangzhou}
  \country{China}}
\email{sijiemai@m.scnu.edu.cn}

\author{Ying Zeng}
\orcid{0000-0001-8842-2045}
\affiliation{%
  \institution{Sun Yat-Sen University}
  \city{Guangzhou}
  \country{China}}
\email{zengy268@mail2.sysu.edu.cn}

\author{Qiaolin He}
\orcid{0009-0001-2204-8668}
\affiliation{%
  \institution{Sun Yat-Sen University}
  \city{Guangzhou}
  \country{China}}
\email{heqlin5@mail2.sysu.edu.cn}

\author{Aolin Xiong}
\orcid{0009-0005-2301-7897}
\affiliation{%
  \institution{Sun Yat-Sen University}
  \city{Guangzhou}
  \country{China}}
\email{xiongaolin@mail2.sysu.edu.cn}

\author{Haifeng Hu}
\orcid{0000-0002-4884-323X}
\affiliation{%
  \institution{Sun Yat-Sen University}
  \city{Guangzhou}
  \country{China}}
\email{huhaif@mail.sysu.edu.cn}


\renewcommand{\shortauthors}{Ronghao Lin et al.}

\begin{abstract}
 This paper presents the winning approach for the $1^{st}$ MultiModal Deception Detection (MMDD) Challenge at the $1^{st}$ Workshop on Subtle Visual Computing (SVC). Aiming at the domain shift issue across source and target domains, we propose a Multi-source Multimodal Progressive Domain Adaptation (MMPDA) framework that transfers the audio-visual knowledge from diverse source domains to the target domain. By gradually aligning source and the target domain at both feature and decision levels, our method bridges domain shifts across diverse multimodal datasets. Extensive experiments demonstrate the effectiveness of our approach securing Top-2 place. Our approach reaches $60.43\%$ on accuracy and $56.99\%$ on F1-score on competition stage 2, surpassing the $1^{st}$ place team by $5.59\%$ on F1-score and the $3^{rd}$ place teams by $6.75\%$ on accuracy. Our code is available at \url{https://github.com/RH-Lin/MMPDA}.



\end{abstract}




\maketitle

\section{Introduction}
\label{sec:introduction}
Audio-visual deception detection aims at predicting the deception state of human by integrating the multimodal information derived from both speech and facial movement \cite{guo2024benchmarking}. Given inherent heterogeneity between audio and visual modalities, prior research focuses on bridging the modality gap by developing sophisticated multimodal fusion strategies at both the feature and decision levels \cite{ding2019face,kumar2021deception,yang2021constructing,karnati2022lienet}. While they have achieved promising results on controlled laboratory benchmarks, they often suffer significant performance degradation when deployed in real-world settings \cite{belavadi2020multimodal}. Specifically, models trained on constrained scenarios tend to generalize poorly to spontaneous interactions confronted with natural variation in talking style, experiment conditions, or background noise. These challenges underscore the urgent need for larger and more diverse datasets to improve the performance and robustness in deception detection \cite{sen2022multimodal}.

\begin{figure}[htbp]
    \vspace{-0.2cm}
    \centering 
    \subfloat[Audio Spectrogram] {
     \label{data_vis_audio}
    \includegraphics[width=1.0\columnwidth]{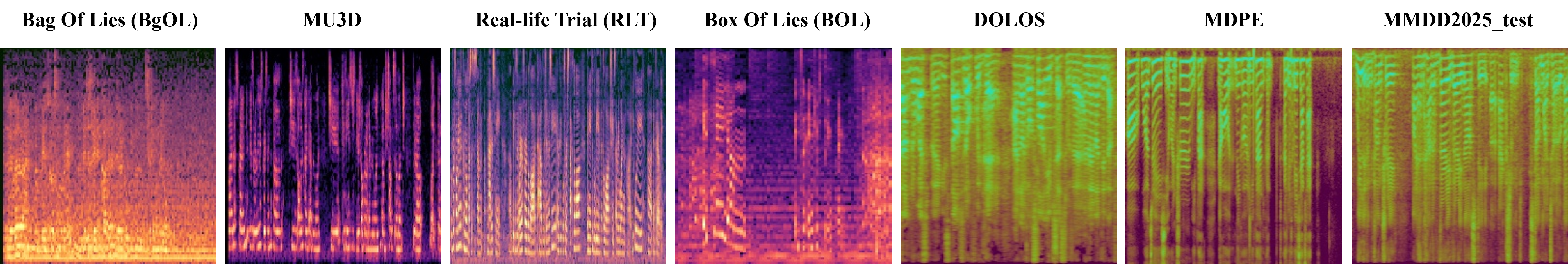}
    } 
    \\
    \vspace{-0.2cm}
    \subfloat[Visual Face Frame] { 
    \label{data_vis_face}
    \includegraphics[width=1.0\columnwidth]{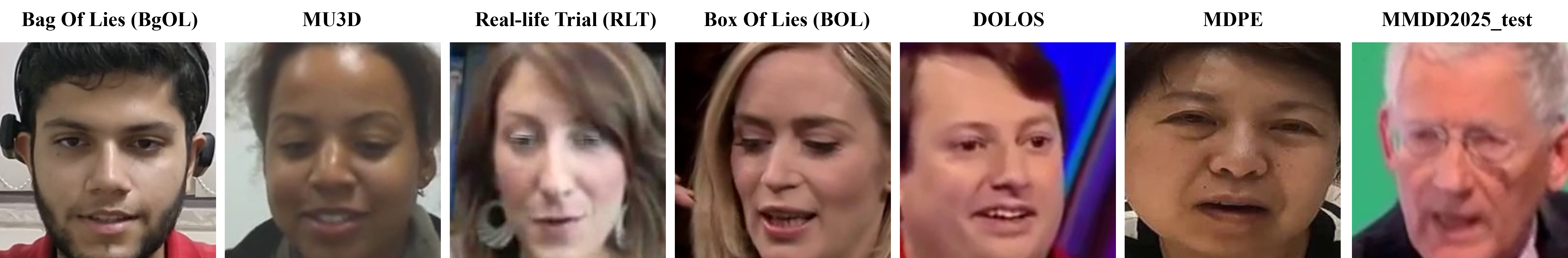}     
    } 
    \vspace{-0.2cm}
    \caption{Visualization of (a) audio spectrogram data and (b) visual face frame data cases for different dataset domains.}
    \label{vis_datasets}
    \vspace{-0.2cm}
\end{figure}

However, the development of generalizable deception detection systems is impeded by two main issues. First, due to the difficulty of collecting and annotating deception data, existing multimodal deception datasets are often limited in the number of subjects and samples \cite{cai2024mdpe}. This scarcity restricts the ability of models to generalize beyond the training distribution.  Second, the available datasets are collected in widely varying scenarios, including courtroom proceedings, laboratory experiments, and game shows \cite{gupta2019bagoflies,lloyd2019mu3d,perezrosas2015rlt,soldner2019boxoflies}, leading to substantial variation for both audio and visual data. 

As shown in Figure \ref{vis_datasets}, the discrepancies across datasets manifest in numerous factors: pitch, tone, loudness, and background noise in the audio modality, as well as resolution, lighting conditions, facial pose, and racial diversity in the visual modality \cite{guo2024benchmarking}. These domain-specific variations greatly impact model performance and highlight the need for robust methods capable of mitigating the effects of both modality gap and domain shift \cite{zhao2020multi,li2024domainadaption,dong2025mmdasurvey}. 

To tackle these issues, we introduce a novel framework named Multi‑source Multimodal Progressive Domain Adaptation (MMPDA) that efficiently leverages complementary multimodal information from diverse source domains, considering the covariance correlation and density divergence across diverse domains and prediction entropy inside each domain. Besides, we adopt multimodal domain adversarial learning to enhance cross-domain alignment with the assistance of gradient reversal backpropagation strategy. Moreover, we strengthen the detection performance through training strategies with domain-by-domain settings and multi-task learning across diverse modalities. Lastly, our framework demands significant improvement across six deception detection datasets, illustrating the generalization and universality ability of the proposed MMPDA. Our key contributions are summarized as follows:

\begin{itemize}
    \item We visualize the domain shift for each unimodal features across the source and target domains, highlighting the motivation of the demands of multimodal domain adaption. 
    \item By gradually aligning multimodal features across each source domain and the target one at both feature and decision levels, our approach effectively minimizes the negative effects caused by modality gap and domain shift.
    \item Extensive experiments validate the effectiveness of the proposed approach on six multimodal deception detection datasets, achieving the state-of-the-art accuracy performance with $63.37\%$ on Stage 1 and $60.43\%$ on Stage 2 at the $1^{st}$ MultiModal Deception Detection (MMDD) Challenge.
\end{itemize}

\section{Related Work}
\label{sec:relatedwork}

\subsection{Multimodal Deception Detection}
Deception is inherently a communicative act grounded in human intent and semantic content. Early work focused largely on text-based signals, leveraging linguistic features such as n-grams, syntactics, readability, and lexical complexity to flag deceptive writing styles \cite{perez2015experiments}. However, deception is more like a multimodal phenomenon, meaning that cues to deception arise not only in the words we choose, but also in how we speak and act. To capture richer signals, more recent studies have turned to multimodal frameworks that integrate different modalities to conduct more accurate deception prediction \cite{karimi2018toward,belavadi2020multimodal,sen2022multimodal}. Among multimodal input data, non-verbal data including audio and visual modalities have been validated as important as verbal data in deception detection \cite{perez2015verbal}. Specifically, audio features including Mel‑frequency cepstral coefficients (MFCCs), pitch, tone, speech polarity, reveal the subtle hesitation patterns characteristic of deceptive speech while visual modalities including facial micro‑expressions, eye blink dynamics, and head and gaze movements, has long linked with deception empirically in physiological responses \cite{vance2022deception}. 

To fuse the heterogeneous audio-visual data, researchers have explored both feature‐level and decision‐level strategies \cite{gogate2017deep,guo2024benchmarking}. Feature‐level fusion concatenates and interacts intermediate representations from unimodal encoders, aiming at learning cross‐modal dynamics beneficial for deception perception \cite{ding2019face,kumar2021deception}. By contrast, decision-level fusion aggregates the outputs of separately trained classifiers for each modality by weighted averaging or meta‐learners to make the model focus on extracting powerful marginal representations from each modality while reducing the computational complexity \cite{yang2021constructing,karnati2022lienet}. Both feature- and decision-level fusion approaches substantially enhance the performance for deception detection system.

Massive deception datasets have been collected through diverse sensors and experiment settings \cite{gupta2019bagoflies,lloyd2019mu3d,perezrosas2015rlt,soldner2019boxoflies}. However, due to the hard evaluation and strict annotation process of deception detection \cite{vance2022deception}, most datasets are typically with relatively small size, which has increasingly limited the further progress of deception detection field \cite{guo2023dolos,cai2024mdpe}. Besides, although the multimodal systems trained on specific dataset may reach appreciable gains on laboratory benchmark, they remain vulnerable to cross-domain shifts when adopted in other datasets or in the real world scenarios \cite{guo2024benchmarking}. Thus, addressing such domain generalization issue is critical for deploying deception detection on the combination of diverse dataset for higher performance and robustness in practical application.

\subsection{Multi-source Domain Adaption}
Domain Adaption, an essential branch of transfer learning, has been studied for improving the generalization ability of model by enabling cross-domain knowledge transfer from the source domain to the target domain \cite{li2024domainadaption}. Previous methods have developed various adaption strategies including feature regularization \cite{xu2020adversarial}, meta learning \cite{li2018metalearning}, model ensembling \cite{koyejo2022ensemble} and so on. 

Leveraging data from multiple source domains \cite{zhang2015multisource,zhao2020multi} to capture the most task-shared information has been proved beneficial for performance on the target relevant task empirically \cite{zhao2024more}. Therefore, multi-source domain adaption are proposed to address the distinct distribution characteristics of each source domain with the target one jointly. Moreover, lots of domain adaption strategies are designed for specialized modality data, such as facial dataset \cite{shi2020towards}, font and handwriting dataset \cite{shankar2018generalizing}, speech and audio dataset \cite{mathur2020libriadapt} and so on. To introduce diverse information from multiple modality streams, multimodal domain adaption has been presented to transferring cross-modal knowledge across different modalities from source to target domain \cite{chen2021mind,cheng2024distengled}. Combining both goals, multi-source multimodal domain adaption approach raises growing attention recently, aiming at harmonizing both domain and modality diversity in a unified framework by aligning multimodal feature distributions across diverse domains \cite{dong2025mmdasurvey,zhao2025mmda}. 

Building on these developments, we introduce a progressive multi‑source multimodal domain adaptation framework that  transfer complementary audio-visual information from source to target domains, which efficiently utilize the deception cues contained in different modalities and datasets.

\section{Methodology}
\label{sec:method}
\begin{figure}[htbp]
	\centering 
	\includegraphics[scale=0.2] {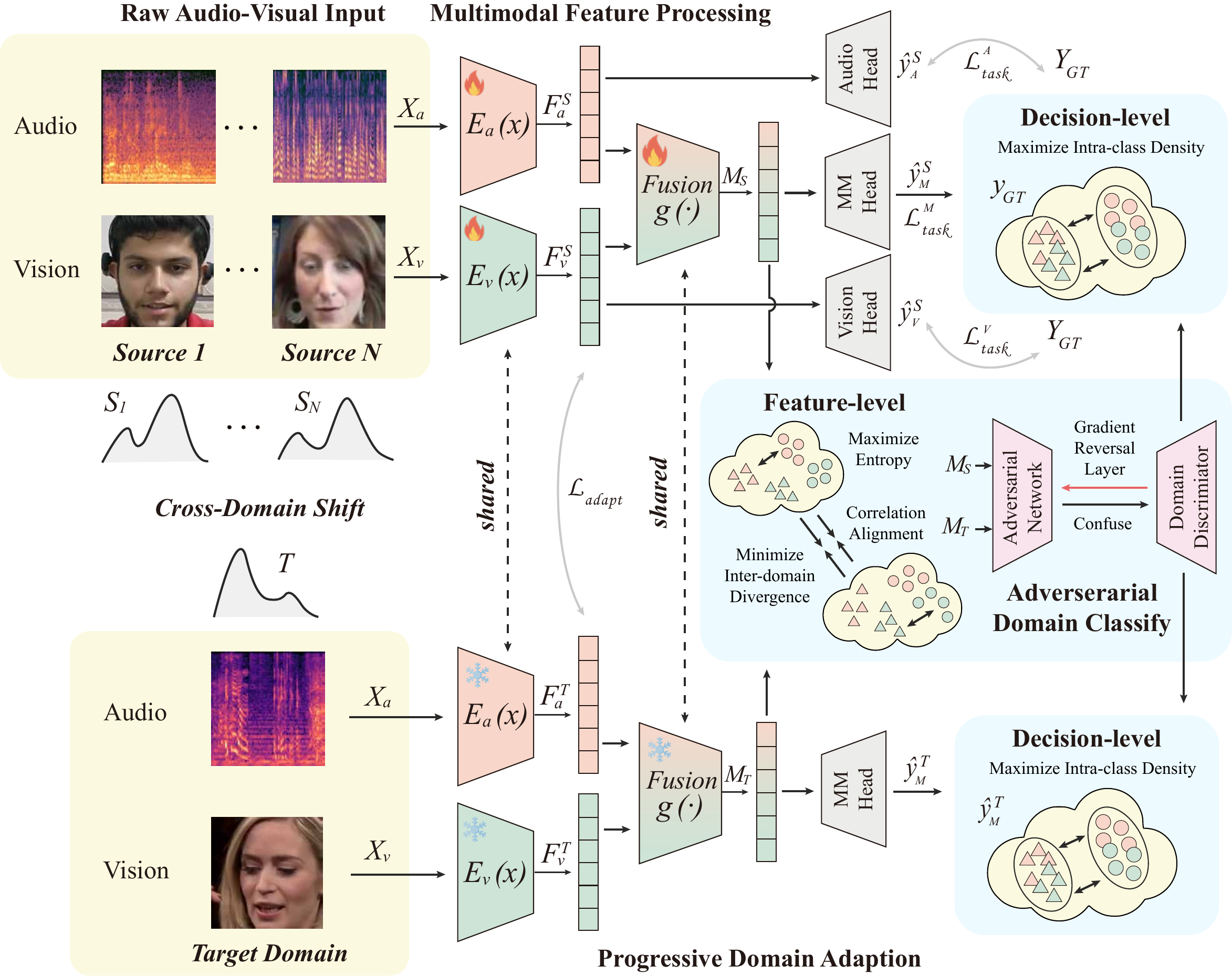}
	\caption{Framework of the proposed MMPDA, consisting of multimodal feature processing and progressive domain adaption. Here $u\in\{a,v\}$ denotes audio and visual modalities.}
	\label{figure0}
    \vspace{-0.3cm}
\end{figure}

This section presents the proposed MMPDA framework as shown in Figure \ref{figure0}. First, we define the problem of multimodal deception detection (Sec. \ref{subsec:definition}) and describe the multimodal feature processing method (Sec. \ref{subsec:feature}). Then, we visualize the cross-domain features to show the domain shift among different deception datasets (Sec. \ref{subsec:gap}). Next, we present our solution by progressively conduct feature- and decision-level multimodal domain adaption (Sec. \ref{subsec:progressive}). Finally, we explain the training strategy to perform the optimal performance for the proposed method (Sec. \ref{subsec:strategy}).

\subsection{Problem Definition}
\label{subsec:definition}
Considering speaker videos with talking speeches and face frames, the goal of audio-visual deception detection is to integrate non-verbal information from acoustic and visual modalities, denoted as $X_{a}$ and $X_{v}$. Extracted by unimodal feature encoders $E(\cdot)$ and integrated by multimodal fusion $g(\cdot)$, we can obtain the final multimodal representations, denoted as:
\begin{equation}
    M=g(E(X_a);E(X_v))=g(F_a;F_v)
\end{equation}

Representing the ground truth label $Y_{gt}$ where $1$ denotes truth while $0$ denotes lie, we can model the deception detection task as a binary classification problem. Obtaining the prediction with multiple linear layers $\hat{y}=MLP(M)$, the Cross-Entropy loss function of deception detection is formulated  as: 
\begin{equation}
    \mathcal{L}_{task} = - [Y_{gt} \log (\hat{y})+ (1-Y_{gt}) \log (1-\hat{y})]
\label{equ_task_form}
\end{equation}

For multi-source domain adaption, there exists multiple source deception domains and the source and target domain are diverse. We denote audio-visual data from source domains $\{S_i\}|_{i=1}^N$ as $S_i=\{(X^S_a,X^S_v;Y_{gt})\}$ and target domains $T=\{(X^T_a,X^T_v)\}$, where $N$ denotes the number of source domain, and $Y_{gt}$ denotes the labels of source domain while target domain is unsupervised during training.

\subsection{Multimodal Feature Processing}
\label{subsec:feature}
Remaining consistent as previous multimodal deception detection methods \cite{guo2024benchmarking}, widely-adopted unimodal encoder tools are employed to extract features from audio and visual data, with each encoder utilizing prior knowledge tailored to modality-specific dynamics. For acoustic modality data, Mel spectrograms extraction is adopted for all audio frame in each video clip, denoted as $x_a\in\mathbf{R}^{224* 224* 3}$. For visual modality data, 32 face frames are sampled for each video clip, denoted as $x_f\in\mathbf{R}^{32*224* 224* 3}$. Moreover, we use ResNet18 \cite{he2016resnet} and LSTM as the backbone network to further extract the audio-visual features $F_a$ and $F_f$ on video frames, whose parameters are trainable in the proposed framework.

Besides, since expressive face act as an essential role in capturing deception intent of human \cite{yang2021constructing}, Emonet \cite{toisoul2021estimation} is utilized to extract 7-dimensional Affect features (5 dimension for expression; 1 dimension for arousal; 1 dimension for valence) and Openface \cite{amos2016openface} is used to extracted the 8-dimensional gaze features and 35-dimensional Action Unit (AU) features. To capture subtle emotion variation, above expressive features extraction are adopted for 64 face frames in each clip, denoted as behavior features $F_b\in\mathbf{R}^{64*50}$. 

In summary, the unimodal features for acoustic and visual modalities are denoted as $X_a \equiv x_a$ and $X_v \equiv \{x_f,x_b\}$. After unimodal extraction, we leverage MulT \cite{tsai2019multimodal} as the multimodal fusion module to conduct cross-modal interaction and learn the final multimodal representation $M=g(E(x_a);E(x_f);E(x_b))=MulT(F_a;F_f;F_b)$.

\subsection{Cross-Domain Shift Visualization}
\label{subsec:gap}
\begin{figure*}[htbp]
    \centering 
    \subfloat[Audio Spectrogram Feature $F_a$] {
     \label{feature_vis_audio_spectrogram}
    \includegraphics[width=0.6\columnwidth]{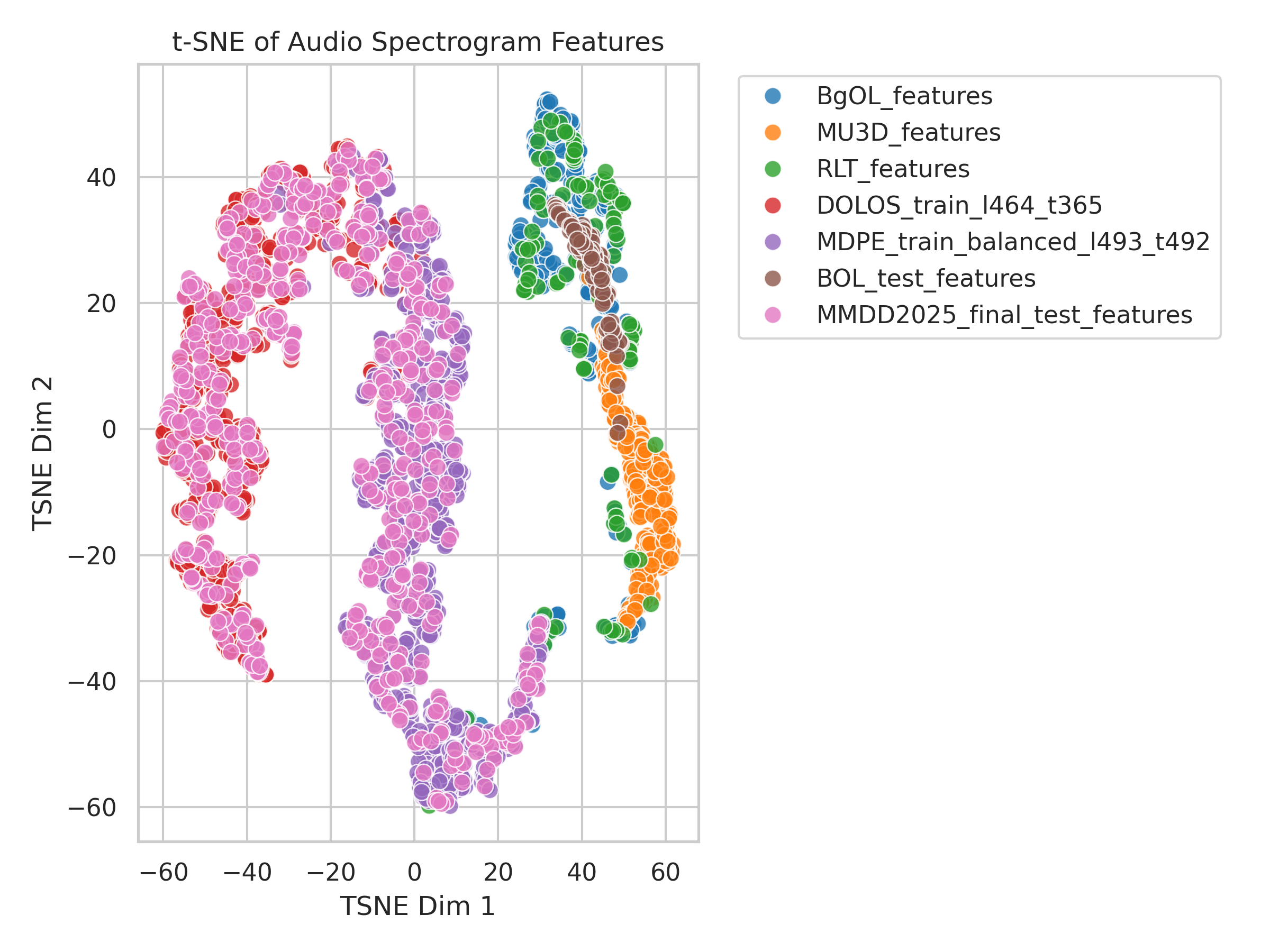}
    } 
    \subfloat[Visual Face Feature $F_f$] { 
    \label{feature_vis_face}
    \includegraphics[width=0.6\columnwidth]{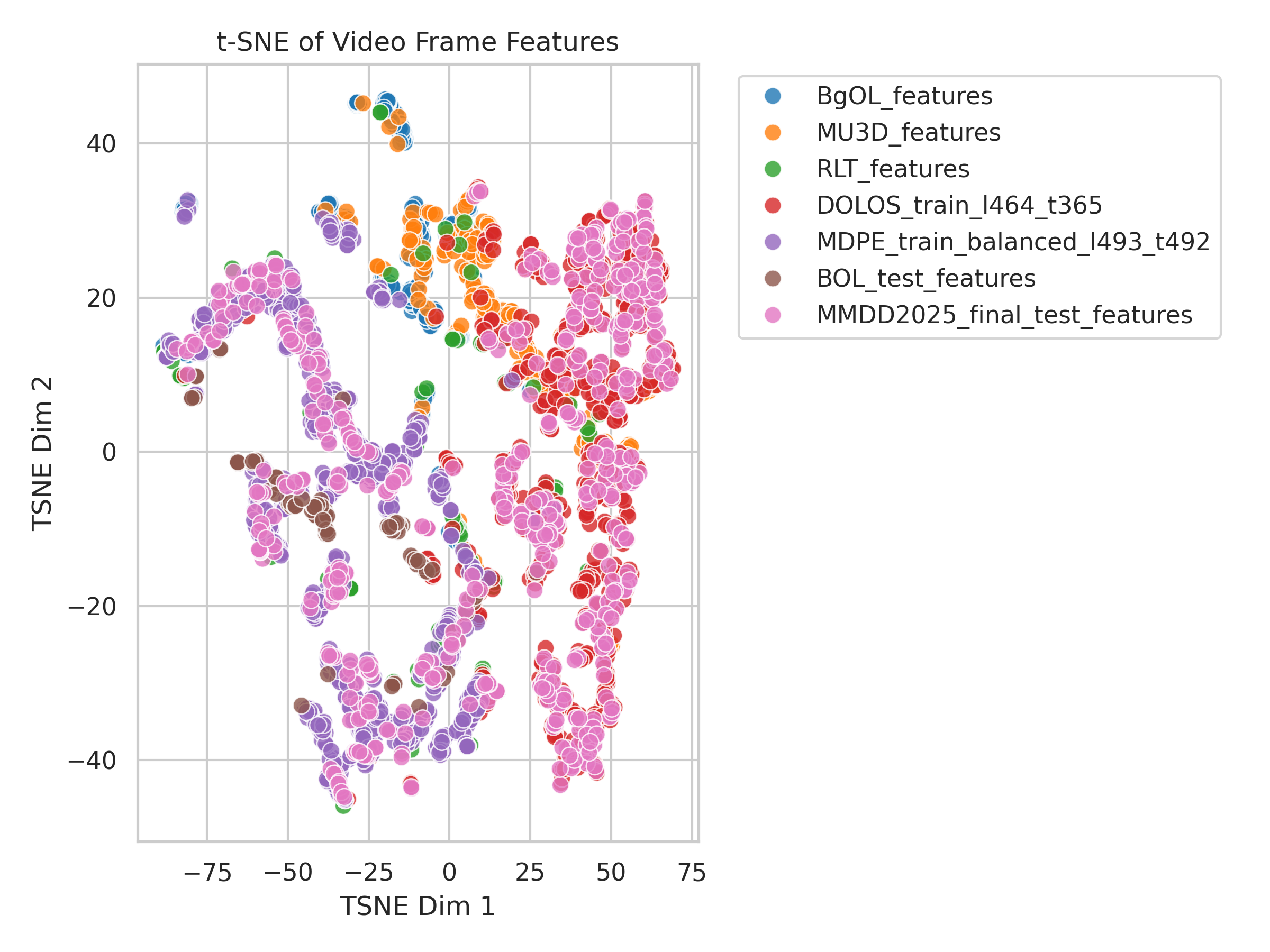}     
    }
    \subfloat[Behavior Feature $F_b$] { 
    \label{feature_vis_behavior}
    \includegraphics[width=0.6\columnwidth]{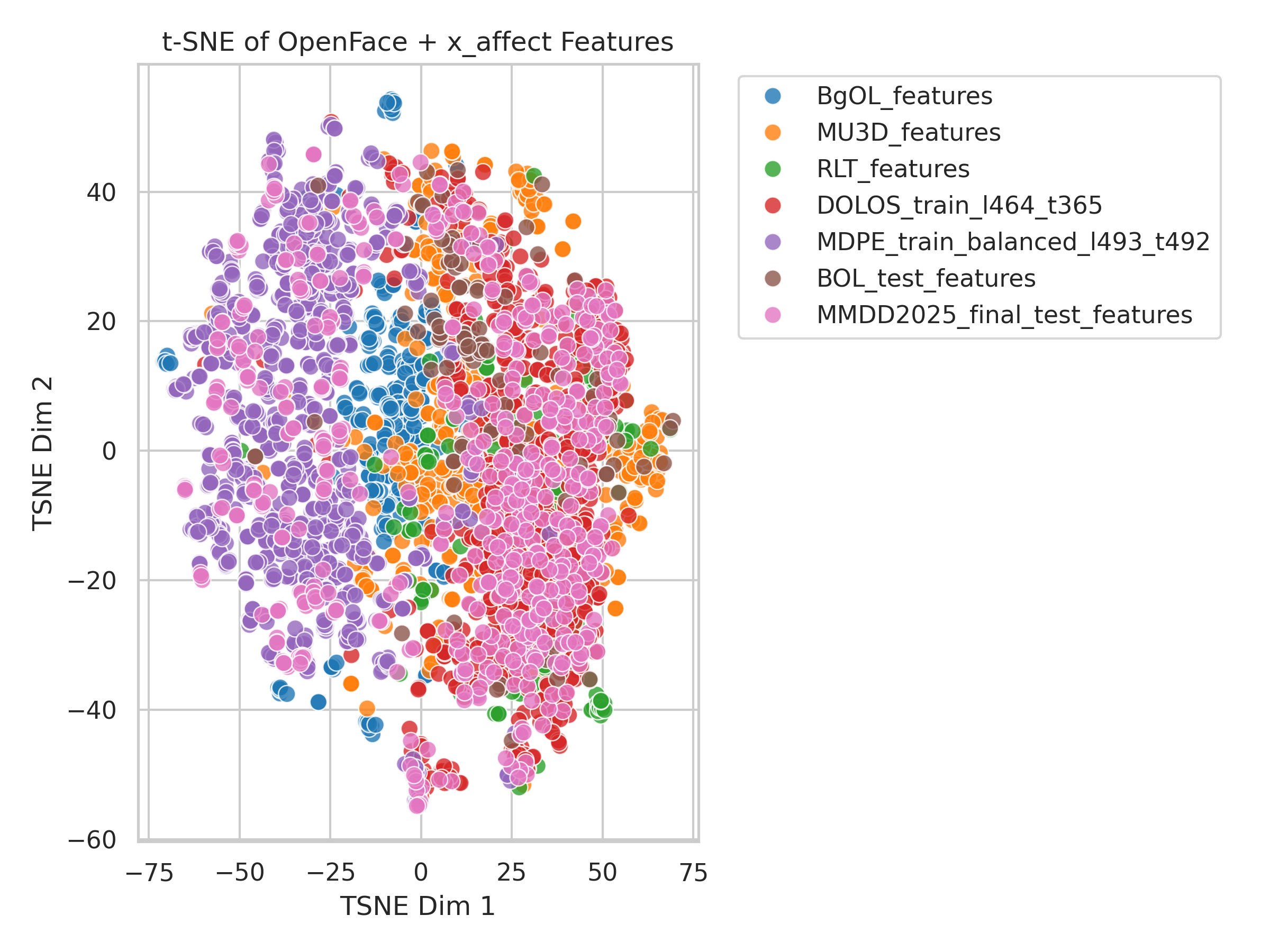}
    }
    \caption{T-SNE visualization for unimodal features in different dataset domains.}
    \label{vis_domain_shift}
    \vspace{-0.1cm}
\end{figure*}

For datasets used in the $1^{st}$ MultiModal Deception Detection (MMDD) Challenge \cite{MMDD2025}, we utilize t-SNE algorithm to visualize the unimodal features for audio-visual input data on each domain as shown in Figure \ref{vis_domain_shift}. The cross-domain shift can be clearly observed on the level of dataset feature domains, no matter for which modality, revealing the motivation and necessity of the proposed multi-source multimodal domain adaption approach.



\subsection{Progressive Domain Adaption}
\label{subsec:progressive}
To tackle the cross-domain shift issue, we sample equal number of multimodal features from source and target domains, denoted as $M_S$ and $M_T$, which are obtained after conducting multimodal fusion on the unimodal features $F_u$ where $u\in\{a,v\}$. Note that for target domains, the network parameters of extracting multimodal features are frozen to avoid training collapse. Then, we present progressive domain adaption for the multimodal feature at both feature- and decision-level consisting of four losses, computed as:
\begin{equation}
    \mathcal{L}_{adapt} = \alpha\mathcal{L}_{CORAL} + \beta\mathcal{L}_{MDD} + \gamma\mathcal{L}_{Entropy} + \eta\mathcal{L}_{Adv}
\label{equ_adapt}
\end{equation}
where $\alpha$, $\beta$, $\gamma$ and $\eta$ are the hyper-parameters to adjust the adaption strength of diverse domain adaption losses.

\subsubsection{\textbf{Deep Correlation Alignment}}
Considering the second-order statistics with the covariance distance of the feature distribution, we adopt Deep Correlation Alignment \cite{sun2016coral} between source and target multimodal features $M_S$ and $M_T$, formulated as:
\begin{equation}
    \mathcal{L}_{CORAL}=\frac{1}{4d^2}\sum_{M_S, M_T} \parallel C_S -C_T \parallel^2 
\end{equation}
where $d$ denotes the feature dimension and the covariance matrices $C$ of source and target features are computed as:
\begin{equation}
\begin{aligned}
    C_S &= \frac{1}{n}(M_S^\top M_S-\frac{1}{n}(\mathbf{1}^\top M_S)^\top(\mathbf{1}^\top M_S))\\
    C_T &= \frac{1}{n}(M_T^\top M_T-\frac{1}{n}(\mathbf{1}^\top M_T)^\top(\mathbf{1}^\top M_T))\\
\end{aligned}
\end{equation}
where $\mathbf{1}$ is a column vector with all one elements and $n$ denotes the number of samples.

\subsubsection{\textbf{Density Divergence Maximization}} To learn the domain-invariant features across diverse domains and make intra- domain features more compact, we calculate the Maximum inter-domain Density Divergence loss \cite{li2021mdd} as: 

\begin{footnotesize}
\begin{equation}
\begin{aligned}
    \mathcal{L}_{MDD} =& \mathbf{E}_{M_S,M_T} \parallel M_S -M_T \parallel^2 +  \mathbf{E}_{M_S,M'_S} \parallel M_S - M'_S \parallel^2 + \mathbf{E}_{M_T,M'_T} \parallel M_T - M'_T \parallel^2 \\
    =& \frac{1}{n} \sum^n_i \parallel M_S -M_T \parallel^2 + 
    \frac{1}{n_S} \sum^{n_S}_i \parallel M_S - M'_S \parallel^2 + \frac{1}{n_T} \sum^{n_T}_i \parallel M_T - M'_T \parallel^2
\end{aligned}
\end{equation}
\end{footnotesize}

The first term calculate pair-wise divergence at relative position for source and target features where $n$ denotes the batch size of both source and target domains. The second or third terms calculate the source or target features with consistent labels respectively in the same batch by iteratively sampling, where $n_S$ and $n_T$ denote the number of $M\cup M'$ satisfying ground truth labels $Y_{gt}=Y'_{gt}$ for source domain or pseudo labels $\hat{y}^T_M=\hat{y}'^T_M$ for target domain. Both inter-domain divergence and intra-domain density are both improved by considering both marginal and conditional distribution supervised with labels at decision-level.

\subsubsection{\textbf{Multimodal Domain Adversarial Learning}}
Inspired by GANs \cite{goodfellow2014gan} which utilize generator and discriminator as adversarial training to conduct generation task, state-of-the-art domain adaption methods usually employ Domain Adversarial Learning (DAL) to minimize the discrepancy between source and target domain in the feature space \cite{li2021mdd,acuna2022domain}. Therefore, we adopt Multimodal Domain Adversarial Learning on the source and target multimodal features $\{M_S,M_T\}$ following Conditional Adversarial Domain Adaptation (CDAN) \cite{long2018conditional}, formulated as:

\begin{tiny}
\begin{equation}
    \min_{E,g}\max_{D} \mathcal{L}_{Adv} = -\mathbf{E}_{M_S}[\sum^C_{c=1}\mathbf{1}_{[c]}\log\sigma(M_S)]+(\mathbf{E}_{M_S}[\log D(h_S)]+\mathbf{E}_{M_T}[\log (1-D(h_T))])
\end{equation}
\end{tiny}
where the first term is a supervised cross-entropy loss to conduct deception detection on the source domain (calculated as the $\mathcal{L}_{task}$ in our paper) and the second term is a conditional loss to conduct domain classification on $h=\Pi(M,\hat{y}_M)$ considering the multimodal features $M$ and its corresponding predictions $\hat{y}_M$. The conditional adversarial network $\Pi$ extracts domain-specific cues and the domain discriminator $D$ determines whether the feature belongs to the source or target domain. 

We adopt entropy condition by reweighting each samples $M$ with an entropy-aware weight $w(\sigma(\hat{y}_M))=1+e^{-\sigma(\hat{y}_M)}$ in our experiments, which is validated for improved transferability in previous works \cite{long2018conditional,li2021mdd}. By incorporating the entropy minimization principle \cite{grandvalet2004semisupervised} in the decision level, the domain discriminator promotes confident and certain predictions, thereby enabling the multimodal adversarial learning to conduct semi-supervised training on unlabeled target domain.

\subsubsection{\textbf{Intra-domain Entropy Maximization}} Previous methods require entropy minimization to learn discrimnative representation. In contrast, we reverse the process into entropy maximization for features $M$ inside both source and target domain to further confuse the domain discriminator \cite{li2021mdd} and enhance the performance of domain adaption, formulated as:
\begin{equation}
     \mathcal{L}_{Entropy} = \frac{1}{n}\sum_{i=1}^n\sigma(M_S;M_T)\log\sigma(M_S;M_T)
\end{equation}

\subsection{Training Strategy}
\label{subsec:strategy}
\subsubsection{\textbf{Gradient Reversal Backpropagation}}
Since there are conflict demands of feature extraction for the original task prediction and the domain discriminator, we introduce Gradient Reversal Layer to flips the sign of the gradient during backward, which has profound implications on the training dynamics and asymptotic learning stability \cite{ganin2015grl}. By inserting gradient reversal operation into the output features $M$ of multimodal fusion module, the training process transforms gradient descent into a competitive pattern, formulated as:
\begin{equation}
     \theta_{E,g}\leftarrow \theta_{E,g}-\mu[\frac{\partial\mathcal{L}_{task}}{\partial\theta_{E,g}}-(\frac{2}{1+e^{-10p}}-1)\cdot\frac{\partial\mathcal{L}_{Adv}}{\partial\theta_{E,g}}]
\end{equation}
where $\theta_{E,g}$ denotes the parameters in unimodal feature encoders $E_u$ and multimodal fusion module $g(\cdot)$ and $p\in(0,1]$ denotes current training progress.

\subsubsection{\textbf{Domain-by-domain Training}}
We train the model by feeding data from multiple source domains one by one, named domain-by-domain training \cite{guo2024benchmarking}, which enable the model to capture domain-specific variations and understand each domain individually before generalizing across diverse domains.

\subsubsection{\textbf{Multi-task Training}}
We conduct multi-task training on each unimodal features to effectively imptove multiple task through mutual learning, which has shown better performance in various multimodal tasks \cite{guo2023dolos}. For $\{F_a/F_v\equiv F_f,F_b\}$, we conduct deception detection on each unimodal features jointly with multimodal features $M$ in the form of Equ. \ref{equ_task_form}, computed as:
\begin{equation}
    \mathcal{L}_{task} = \sum_{u\in\{a,v\}}\mathcal{L}^u_{task} + \mathcal{L}^M_{task}
\label{equ_task}
\end{equation}

\subsubsection{\textbf{Optimization Objective}}
In general, combining the task prediction loss in Equ. \ref{equ_task} and the domain adpation loss in Equ. \ref{equ_adapt}, the total optimization objective is formulated as:
\begin{equation}
\label{equ_total}
    \mathcal{L}_{total} = \mathcal{L}_{task} + \lambda\mathcal{L}_{adapt}
\end{equation}

\section{Experiments}

\subsection{Datasets and Evaluation Metrics}
Six multimodal deception detection datasets are used in the $1^{st}$ MMDD Competition, including including Bag Of Lies (BgOL) \cite{gupta2019bagoflies}, MU3D \cite{lloyd2019mu3d}, Real-life Trial (RLT) \cite{perezrosas2015rlt}, DOLOS \cite{guo2023dolos}, MDPE \cite{cai2024mdpe} as the source domains $S_i$ while Box Of Lies (BOL) \cite{soldner2019boxoflies} as the target domain in competition stage 1 and MMDD 2025 testing subset datasets \cite{MMDD2025} which is sampled from DOLOS and MDPE testing subset as the target domain in competition stage 2. Following previous methods, we utilize binary classification accuracy and F1-score (expressed as a percentage value) as the evaluation metric for our experiments.

\subsection{$1^{st}$ MMDD Competition Stage 1 Result}
Competition Stage 1 utilizes BgOL, MU3D or RLT as the source domain and BOL as the target domain. As shown in Table \ref{exp_stage1}, performance improvement for single-to-single domain adaption with diverse source domains indicates the effectiveness of MMPDA in cross-domain generalization by aligning multimodal features.

\begin{table}[htbp]
\caption{Performance comparison between the baseline and the proposed MMPDA in Competition Stage 1. Here binary accuracy is used as evaluation metric.}
\label{exp_stage1}
\centering
\begin{tabular}{c|c|c|c}
    \hline
    Models & BgOL$\rightarrow$BOL & MU3D$\rightarrow$BOL & RLT$\rightarrow$BOL \\
    \hline
    Baseline & 57.43 & 60.40 & 55.45 \\
    \textbf{MMPDA} & 58.42 \textcolor[rgb]{0.15, 0.65, 0.15}{(+0.99)} & 63.37 \textcolor[rgb]{0.15, 0.65, 0.15}{(+2.97)} & 63.37 \textcolor[rgb]{0.15, 0.65, 0.15}{(+7.92)} \\
    \hline
\end{tabular}
\vspace{-0.2cm}
\end{table} 

\subsection{$1^{st}$ MMDD Competition Stage 2 Result}
Competition Stage 2 utilizes DOLOS or MDPE as the source domains and MMDD 2025 testing subset as the target domain. As shown in Table \ref{exp_stage2}, MMPDA consistently outperforms the baseline by exploiting the domain-invariant knowledge. Combining multiple source domains, MMPDA delivers stronger improvements, showing its productive multi-source information integration capability.

\begin{table}[htbp]
\vspace{-0.1cm}
\caption{Performance comparison between the baseline and the proposed MMPDA in Competition Stage 2.}
\label{exp_stage2}
\centering
\begin{tabular}{c|c|c|c}
    \hline
    Source & Models & Accuracy & F1-score \\
    \hline
    \multirow{2}{*}{DOLOS} & Baseline & 51.75 & 44.92 \\
    & \textbf{MMPDA} & 57.43 \textcolor[rgb]{0.15, 0.65, 0.15}{(+5.68)} & 51.43 \textcolor[rgb]{0.15, 0.65, 0.15}{(+6.51)} \\
    \hline
    \multirow{2}{*}{MDPE}  & Baseline & 50.93 & 46.43 \\
    & \textbf{MMPDA} & 58.93 \textcolor[rgb]{0.15, 0.65, 0.15}{(+8.00)} & 56.34 \textcolor[rgb]{0.15, 0.65, 0.15}{(+9.91)} \\
    \hline
    \multirow{2}{*}{DOLOS, MDPE} & Baseline & 54.92 & 49.13 \\
    & \textbf{MMPDA} & 60.43 \textcolor[rgb]{0.15, 0.65, 0.15}{(+5.51)} & 56.99 \textcolor[rgb]{0.15, 0.65, 0.15}{(+7.86)} \\
    \hline
\end{tabular}
\vspace{-0.2cm}
\end{table} 

\subsection{Ablation Study}
We conduct ablation study on multimodal deception detection with DOLOS and MDPE as the source domains and MMDD 2025 testing subset as the target domain. 
Table \ref{exp_param_sensitivity} examines the impact of varying the values of the hyper-parameters $\alpha$, $\beta$, $\gamma$ and $\eta$ in Equ. \ref{equ_adapt} under different values of $\lambda$ in Equ. \ref{equ_total}, adjusting the adaptation effect for different components. With $\lambda>0$, carefully balancing the contribution of individual loss components (with higher weights on correlation alignment and entropy maximization, and smaller ones on density divergence maximization and adversarial learning) yields optimal performance.

Moreover, Table \ref{exp_ablation_grl} evaluates the importance of the gradient reversal strategy, which enhances the stability of multimodal domain adversarial learning for capturing domain-invariant features and brings a performance boost on accuracy.


\begin{table}[htbp]
\vspace{-0.1cm}
\caption{Parameter sensitivity for $\alpha$, $\beta$, $\gamma$, $\eta$, $\lambda$ of MMPDA.}
\label{exp_param_sensitivity}
\centering
\begin{tabular}{c|cccc|c}
    \hline
    Parameter $\lambda$ & $\alpha$  & $\beta$  & $\gamma$  & $\eta$ & Accuracy \\
    \hline
    0 & - & - & - & - & 54.92 \\
    \hline
    \multirow{10}{*}{1} & 1 & 1 & 1 & 1 & 58.10 \\
    & 0.1 & 1 & 1 & 1 & 56.34 \\
    & 10 & 1 & 1 & 1 & 59.22 \\
    & 1 & 0.1 & 1 & 1 & 58.64 \\
    & 1 & 10 & 1 & 1 & 54.91 \\
    & 1 & 1 & 0.1 & 1 & 56.63 \\
    & 1 & 1 & 10 & 1 & 58.36 \\
    & 1 & 1 & 1 & 0.1 & 58.64 \\
    & 1 & 1 & 1 & 10 & 54.91 \\
    & 10 & 0.1 & 10 & 0.1 & 59.88 \\
    \hline
    10 & 10 & 0.1 & 10 & 0.1 & 60.43 \\
    \hline
\end{tabular}
\vspace{-0.2cm}
\end{table} 


\begin{table}[htbp]
\vspace{-0.2cm}
\caption{Ablation for gradient reversal layer of MMPDA.}
\label{exp_ablation_grl}
\centering
\begin{tabular}{c|c}
    \hline
    Module & Accuracy \\
    \hline
    w/o Gradient Reversal & 58.62 \\
    w Gradient Reversal & 60.43 \\
    \hline
\end{tabular}
\vspace{-0.3cm}
\end{table} 



\section{Conclusion}
In this paper, we proposed our winning approach named Multi-source Multimodal Progressive Domain Adaptation (MMPDA) which reaches Top-2 performance in the $1^{st}$ MultiModal Deception Detection (MMDD) Challenge. To address the challenges of domain shift and modality gap in audio-visual deception detection,
our approach effectively enhances cross-domain adaption by progressively transferring multimodal knowledge from multiple source domains to target on at the levels of feature and decision. Extensive experiments on six public deception datasets demonstrate the effectiveness of our method, highlighting the potential of multi-source multimodal adaptation in building generalizable deception detection systems for real-world applications.


\begin{acks}
This work was supported by the National Natural Science Foundation of China (62076262, 61673402, 61273270, 60802069) and by the International Program Fund for Young Talent Scientific Research People, Sun Yat-Sen University.
\end{acks}

\bibliographystyle{ACM-Reference-Format}
\balance
\bibliography{deception}


\begin{thebibliography}{45}


\ifx \showCODEN    \undefined \def \showCODEN     #1{\unskip}     \fi
\ifx \showISBNx    \undefined \def \showISBNx     #1{\unskip}     \fi
\ifx \showISBNxiii \undefined \def \showISBNxiii  #1{\unskip}     \fi
\ifx \showISSN     \undefined \def \showISSN      #1{\unskip}     \fi
\ifx \showLCCN     \undefined \def \showLCCN      #1{\unskip}     \fi
\ifx \shownote     \undefined \def \shownote      #1{#1}          \fi
\ifx \showarticletitle \undefined \def \showarticletitle #1{#1}   \fi
\ifx \showURL      \undefined \def \showURL       {\relax}        \fi
\providecommand\bibfield[2]{#2}
\providecommand\bibinfo[2]{#2}
\providecommand\natexlab[1]{#1}
\providecommand\showeprint[2][]{arXiv:#2}

\bibitem[Acuna et~al\mbox{.}(2022)]%
        {acuna2022domain}
\bibfield{author}{\bibinfo{person}{David Acuna}, \bibinfo{person}{Marc~T Law},
  \bibinfo{person}{Guojun Zhang}, {and} \bibinfo{person}{Sanja Fidler}.}
  \bibinfo{year}{2022}\natexlab{}.
\newblock \showarticletitle{Domain Adversarial Training: A Game Perspective}.
  In \bibinfo{booktitle}{\emph{International Conference on Learning
  Representations}}.
\newblock


\bibitem[Amos et~al\mbox{.}(2016)]%
        {amos2016openface}
\bibfield{author}{\bibinfo{person}{Brandon Amos}, \bibinfo{person}{Bartosz
  Ludwiczuk}, {and} \bibinfo{person}{Mahadev Satyanarayanan}.}
  \bibinfo{year}{2016}\natexlab{}.
\newblock \bibinfo{booktitle}{\emph{OpenFace: A general-purpose face
  recognition library with mobile applications}}.
\newblock \bibinfo{type}{{T}echnical {R}eport}.
  \bibinfo{institution}{CMU-CS-16-118, CMU School of Computer Science}.
\newblock


\bibitem[Arpit et~al\mbox{.}(2022)]%
        {koyejo2022ensemble}
\bibfield{author}{\bibinfo{person}{Devansh Arpit}, \bibinfo{person}{Huan Wang},
  \bibinfo{person}{Yingbo Zhou}, {and} \bibinfo{person}{Caiming Xiong}.}
  \bibinfo{year}{2022}\natexlab{}.
\newblock \showarticletitle{Ensemble of Averages: Improving Model Selection and
  Boosting Performance in Domain Generalization}. In
  \bibinfo{booktitle}{\emph{Advances in Neural Information Processing
  Systems}}, Vol.~\bibinfo{volume}{35}. \bibinfo{publisher}{Curran Associates,
  Inc.}, \bibinfo{pages}{8265--8277}.
\newblock


\bibitem[Belavadi et~al\mbox{.}(2020)]%
        {belavadi2020multimodal}
\bibfield{author}{\bibinfo{person}{Vibha Belavadi}, \bibinfo{person}{Yan Zhou},
  \bibinfo{person}{Jonathan~Z. Bakdash}, \bibinfo{person}{Murat Kantarcioglu},
  \bibinfo{person}{Daniel~C. Krawczyk}, \bibinfo{person}{Linda Nguyen},
  \bibinfo{person}{Jelena Rakic}, {and} \bibinfo{person}{Bhavani
  Thuriasingham}.} \bibinfo{year}{2020}\natexlab{}.
\newblock \showarticletitle{MultiModal Deception Detection: Accuracy,
  Applicability and Generalizability}. In \bibinfo{booktitle}{\emph{2020 Second
  IEEE International Conference on Trust, Privacy and Security in Intelligent
  Systems and Applications (TPS-ISA)}}. \bibinfo{pages}{99--106}.
\newblock


\bibitem[Cai et~al\mbox{.}(2024)]%
        {cai2024mdpe}
\bibfield{author}{\bibinfo{person}{Cong Cai}, \bibinfo{person}{Shan Liang},
  \bibinfo{person}{Xuefei Liu}, \bibinfo{person}{Kang Zhu},
  \bibinfo{person}{Zhengqi Wen}, \bibinfo{person}{Jianhua Tao},
  \bibinfo{person}{Heng Xie}, \bibinfo{person}{Jizhou Cui},
  \bibinfo{person}{Yiming Ma}, \bibinfo{person}{Zhenhua Cheng},
  {et~al\mbox{.}}} \bibinfo{year}{2024}\natexlab{}.
\newblock \showarticletitle{MDPE: A Multimodal Deception Dataset with
  Personality and Emotional Characteristics}.
\newblock \bibinfo{journal}{\emph{arXiv preprint arXiv:2407.12274}}
  (\bibinfo{year}{2024}).
\newblock


\bibitem[Chen et~al\mbox{.}(2021)]%
        {chen2021mind}
\bibfield{author}{\bibinfo{person}{Qingchao Chen}, \bibinfo{person}{Yang Liu},
  {and} \bibinfo{person}{Samuel Albanie}.} \bibinfo{year}{2021}\natexlab{}.
\newblock \showarticletitle{Mind-the-Gap! Unsupervised Domain Adaptation for
  Text-Video Retrieval}.
\newblock \bibinfo{journal}{\emph{Proceedings of the AAAI Conference on
  Artificial Intelligence}} \bibinfo{volume}{35}, \bibinfo{number}{2}
  (\bibinfo{date}{May} \bibinfo{year}{2021}), \bibinfo{pages}{1072--1080}.
\newblock


\bibitem[Cheng et~al\mbox{.}(2024)]%
        {cheng2024distengled}
\bibfield{author}{\bibinfo{person}{De Cheng}, \bibinfo{person}{Zhipeng Xu},
  \bibinfo{person}{Xinyang Jiang}, \bibinfo{person}{Nannan Wang},
  \bibinfo{person}{Dongsheng Li}, {and} \bibinfo{person}{Xinbo Gao}.}
  \bibinfo{year}{2024}\natexlab{}.
\newblock \showarticletitle{Disentangled Prompt Representation for Domain
  Generalization}. In \bibinfo{booktitle}{\emph{Proceedings of the IEEE/CVF
  Conference on Computer Vision and Pattern Recognition (CVPR)}}.
  \bibinfo{pages}{23595--23604}.
\newblock


\bibitem[Ding et~al\mbox{.}(2019)]%
        {ding2019face}
\bibfield{author}{\bibinfo{person}{Mingyu Ding}, \bibinfo{person}{An Zhao},
  \bibinfo{person}{Zhiwu Lu}, \bibinfo{person}{Tao Xiang}, {and}
  \bibinfo{person}{Ji-Rong Wen}.} \bibinfo{year}{2019}\natexlab{}.
\newblock \showarticletitle{Face-Focused Cross-Stream Network for Deception
  Detection in Videos}. In \bibinfo{booktitle}{\emph{Proceedings of the
  IEEE/CVF Conference on Computer Vision and Pattern Recognition (CVPR)}}.
\newblock


\bibitem[Dong et~al\mbox{.}(2025)]%
        {dong2025mmdasurvey}
\bibfield{author}{\bibinfo{person}{Hao Dong}, \bibinfo{person}{Moru Liu},
  \bibinfo{person}{Kaiyang Zhou}, \bibinfo{person}{Eleni Chatzi},
  \bibinfo{person}{Juho Kannala}, \bibinfo{person}{Cyrill Stachniss}, {and}
  \bibinfo{person}{Olga Fink}.} \bibinfo{year}{2025}\natexlab{}.
\newblock \showarticletitle{Advances in Multimodal Adaptation and
  Generalization: From Traditional Approaches to Foundation Models}.
\newblock \bibinfo{journal}{\emph{arXiv preprint arXiv:2501.18592}}
  (\bibinfo{year}{2025}).
\newblock


\bibitem[Ganin and Lempitsky(2015)]%
        {ganin2015grl}
\bibfield{author}{\bibinfo{person}{Yaroslav Ganin} {and}
  \bibinfo{person}{Victor Lempitsky}.} \bibinfo{year}{2015}\natexlab{}.
\newblock \showarticletitle{Unsupervised domain adaptation by backpropagation}.
  In \bibinfo{booktitle}{\emph{Proceedings of the 32nd International Conference
  on International Conference on Machine Learning - Volume 37}} (Lille, France)
  \emph{(\bibinfo{series}{ICML'15})}. \bibinfo{publisher}{JMLR.org},
  \bibinfo{pages}{1180–1189}.
\newblock


\bibitem[Gogate et~al\mbox{.}(2017)]%
        {gogate2017deep}
\bibfield{author}{\bibinfo{person}{Mandar Gogate}, \bibinfo{person}{Ahsan
  Adeel}, {and} \bibinfo{person}{Amir Hussain}.}
  \bibinfo{year}{2017}\natexlab{}.
\newblock \showarticletitle{Deep learning driven multimodal fusion for
  automated deception detection}. In \bibinfo{booktitle}{\emph{2017 IEEE
  Symposium Series on Computational Intelligence (SSCI)}}.
  \bibinfo{pages}{1--6}.
\newblock


\bibitem[Goodfellow et~al\mbox{.}(2014)]%
        {goodfellow2014gan}
\bibfield{author}{\bibinfo{person}{Ian~J. Goodfellow}, \bibinfo{person}{Jean
  Pouget-Abadie}, \bibinfo{person}{Mehdi Mirza}, \bibinfo{person}{Bing Xu},
  \bibinfo{person}{David Warde-Farley}, \bibinfo{person}{Sherjil Ozair},
  \bibinfo{person}{Aaron Courville}, {and} \bibinfo{person}{Yoshua Bengio}.}
  \bibinfo{year}{2014}\natexlab{}.
\newblock \showarticletitle{Generative adversarial nets}. In
  \bibinfo{booktitle}{\emph{Proceedings of the 28th International Conference on
  Neural Information Processing Systems - Volume 2}} (Montreal, Canada)
  \emph{(\bibinfo{series}{NIPS'14})}. \bibinfo{publisher}{MIT Press},
  \bibinfo{address}{Cambridge, MA, USA}, \bibinfo{pages}{2672–2680}.
\newblock


\bibitem[Grandvalet and Bengio(2004)]%
        {grandvalet2004semisupervised}
\bibfield{author}{\bibinfo{person}{Yves Grandvalet} {and}
  \bibinfo{person}{Yoshua Bengio}.} \bibinfo{year}{2004}\natexlab{}.
\newblock \showarticletitle{Semi-supervised learning by entropy minimization}.
  In \bibinfo{booktitle}{\emph{Proceedings of the 18th International Conference
  on Neural Information Processing Systems}} (Vancouver, British Columbia,
  Canada) \emph{(\bibinfo{series}{NIPS'04})}. \bibinfo{publisher}{MIT Press},
  \bibinfo{address}{Cambridge, MA, USA}, \bibinfo{pages}{529–536}.
\newblock


\bibitem[Guo et~al\mbox{.}(2023)]%
        {guo2023dolos}
\bibfield{author}{\bibinfo{person}{Xiaobao Guo},
  \bibinfo{person}{Nithish~Muthuchamy Selvaraj}, \bibinfo{person}{Zitong Yu},
  \bibinfo{person}{Adams Wai-Kin Kong}, \bibinfo{person}{Bingquan Shen}, {and}
  \bibinfo{person}{Alex Kot}.} \bibinfo{year}{2023}\natexlab{}.
\newblock \showarticletitle{Audio-visual deception detection: Dolos dataset and
  parameter-efficient crossmodal learning}. In
  \bibinfo{booktitle}{\emph{Proceedings of the IEEE/CVF International
  Conference on Computer Vision}}. \bibinfo{pages}{22135--22145}.
\newblock


\bibitem[Guo et~al\mbox{.}(2024)]%
        {guo2024benchmarking}
\bibfield{author}{\bibinfo{person}{Xiaobao Guo}, \bibinfo{person}{Zitong Yu},
  \bibinfo{person}{Nithish~Muthuchamy Selvaraj}, \bibinfo{person}{Bingquan
  Shen}, \bibinfo{person}{Adams Wai-Kin Kong}, {and} \bibinfo{person}{Alex~C
  Kot}.} \bibinfo{year}{2024}\natexlab{}.
\newblock \showarticletitle{Benchmarking Cross-Domain Audio-Visual Deception
  Detection}.
\newblock \bibinfo{journal}{\emph{arXiv preprint arXiv:2405.06995}}
  (\bibinfo{year}{2024}).
\newblock


\bibitem[Gupta et~al\mbox{.}(2019)]%
        {gupta2019bagoflies}
\bibfield{author}{\bibinfo{person}{Viresh Gupta}, \bibinfo{person}{Mohit
  Agarwal}, \bibinfo{person}{Manik Arora}, \bibinfo{person}{Tanmoy
  Chakraborty}, \bibinfo{person}{Richa Singh}, {and} \bibinfo{person}{Mayank
  Vatsa}.} \bibinfo{year}{2019}\natexlab{}.
\newblock \showarticletitle{Bag-of-Lies: A Multimodal Dataset for Deception
  Detection}. In \bibinfo{booktitle}{\emph{2019 IEEE/CVF Conference on Computer
  Vision and Pattern Recognition Workshops (CVPRW)}}. \bibinfo{pages}{83--90}.
\newblock


\bibitem[He et~al\mbox{.}(2016)]%
        {he2016resnet}
\bibfield{author}{\bibinfo{person}{Kaiming He}, \bibinfo{person}{Xiangyu
  Zhang}, \bibinfo{person}{Shaoqing Ren}, {and} \bibinfo{person}{Jian Sun}.}
  \bibinfo{year}{2016}\natexlab{}.
\newblock \showarticletitle{Deep Residual Learning for Image Recognition}. In
  \bibinfo{booktitle}{\emph{2016 IEEE Conference on Computer Vision and Pattern
  Recognition (CVPR)}}. \bibinfo{pages}{770--778}.
\newblock


\bibitem[Karimi et~al\mbox{.}(2018)]%
        {karimi2018toward}
\bibfield{author}{\bibinfo{person}{Hamid Karimi}, \bibinfo{person}{Jiliang
  Tang}, {and} \bibinfo{person}{Yanen Li}.} \bibinfo{year}{2018}\natexlab{}.
\newblock \showarticletitle{Toward End-to-End Deception Detection in Videos}.
  In \bibinfo{booktitle}{\emph{2018 IEEE International Conference on Big Data
  (Big Data)}}.
\newblock


\bibitem[Karnati et~al\mbox{.}(2022)]%
        {karnati2022lienet}
\bibfield{author}{\bibinfo{person}{Mohan Karnati}, \bibinfo{person}{Ayan Seal},
  \bibinfo{person}{Anis Yazidi}, {and} \bibinfo{person}{Ondrej Krejcar}.}
  \bibinfo{year}{2022}\natexlab{}.
\newblock \showarticletitle{LieNet: A Deep Convolution Neural Network Framework
  for Detecting Deception}.
\newblock \bibinfo{journal}{\emph{IEEE Transactions on Cognitive and
  Developmental Systems}} \bibinfo{volume}{14}, \bibinfo{number}{3}
  (\bibinfo{year}{2022}), \bibinfo{pages}{971--984}.
\newblock


\bibitem[Kumar et~al\mbox{.}(2021)]%
        {kumar2021deception}
\bibfield{author}{\bibinfo{person}{Srijan Kumar}, \bibinfo{person}{Chongyang
  Bai}, \bibinfo{person}{V.S. Subrahmanian}, {and} \bibinfo{person}{Jure
  Leskovec}.} \bibinfo{year}{2021}\natexlab{}.
\newblock \showarticletitle{Deception Detection in Group Video Conversations
  using Dynamic Interaction Networks}.
\newblock \bibinfo{journal}{\emph{Proceedings of the International AAAI
  Conference on Web and Social Media}} \bibinfo{volume}{15},
  \bibinfo{number}{1} (\bibinfo{date}{May} \bibinfo{year}{2021}),
  \bibinfo{pages}{339--350}.
\newblock


\bibitem[Li et~al\mbox{.}(2018)]%
        {li2018metalearning}
\bibfield{author}{\bibinfo{person}{Da Li}, \bibinfo{person}{Yongxin Yang},
  \bibinfo{person}{Yi-Zhe Song}, {and} \bibinfo{person}{Timothy Hospedales}.}
  \bibinfo{year}{2018}\natexlab{}.
\newblock \showarticletitle{Learning to Generalize: Meta-Learning for Domain
  Generalization}.
\newblock \bibinfo{journal}{\emph{Proceedings of the AAAI Conference on
  Artificial Intelligence}} \bibinfo{volume}{32}, \bibinfo{number}{1}
  (\bibinfo{date}{Apr.} \bibinfo{year}{2018}).
\newblock


\bibitem[Li et~al\mbox{.}(2021)]%
        {li2021mdd}
\bibfield{author}{\bibinfo{person}{Jingjing Li}, \bibinfo{person}{Erpeng Chen},
  \bibinfo{person}{Zhengming Ding}, \bibinfo{person}{Lei Zhu},
  \bibinfo{person}{Ke Lu}, {and} \bibinfo{person}{Heng~Tao Shen}.}
  \bibinfo{year}{2021}\natexlab{}.
\newblock \showarticletitle{Maximum Density Divergence for Domain Adaptation}.
\newblock \bibinfo{journal}{\emph{IEEE Transactions on Pattern Analysis and
  Machine Intelligence}} \bibinfo{volume}{43}, \bibinfo{number}{11}
  (\bibinfo{year}{2021}), \bibinfo{pages}{3918--3930}.
\newblock


\bibitem[Li et~al\mbox{.}(2024)]%
        {li2024domainadaption}
\bibfield{author}{\bibinfo{person}{Jingjing Li}, \bibinfo{person}{Zhiqi Yu},
  \bibinfo{person}{Zhekai Du}, \bibinfo{person}{Lei Zhu}, {and}
  \bibinfo{person}{Heng~Tao Shen}.} \bibinfo{year}{2024}\natexlab{}.
\newblock \showarticletitle{A Comprehensive Survey on Source-Free Domain
  Adaptation}.
\newblock \bibinfo{journal}{\emph{IEEE Transactions on Pattern Analysis and
  Machine Intelligence}} \bibinfo{volume}{46}, \bibinfo{number}{8}
  (\bibinfo{year}{2024}), \bibinfo{pages}{5743--5762}.
\newblock


\bibitem[Lin et~al\mbox{.}(2025)]%
        {MMDD2025}
\bibfield{author}{\bibinfo{person}{Xun Lin}, \bibinfo{person}{Xiaobao Guo},
  \bibinfo{person}{Taorui Wang}, \bibinfo{person}{Yingjie Ma},
  \bibinfo{person}{Tiajian Huang}, \bibinfo{person}{Jiayu Zhang},
  \bibinfo{person}{Junzhe Cao}, {and} \bibinfo{person}{Zitong Yu}.}
  \bibinfo{year}{2025}\natexlab{}.
\newblock \showarticletitle{SVC 2025: the first Multimodal deception detection
  Challenge} \emph{(\bibinfo{series}{MM '25 SVC Workshop})}.
  \bibinfo{publisher}{Association for Computing Machinery}.
\newblock
\urldef\tempurl%
\url{https://sites.google.com/view/svc-mm25}
\showURL{%
\tempurl}


\bibitem[Lloyd et~al\mbox{.}(2019)]%
        {lloyd2019mu3d}
\bibfield{author}{\bibinfo{person}{E~Paige Lloyd}, \bibinfo{person}{Jason~C
  Deska}, \bibinfo{person}{Kurt Hugenberg}, \bibinfo{person}{Allen~R
  McConnell}, \bibinfo{person}{Brandon~T Humphrey}, {and}
  \bibinfo{person}{Jonathan~W Kunstman}.} \bibinfo{year}{2019}\natexlab{}.
\newblock \showarticletitle{Miami University deception detection database}.
\newblock \bibinfo{journal}{\emph{Behavior research methods}}
  \bibinfo{volume}{51}, \bibinfo{number}{1} (\bibinfo{date}{February}
  \bibinfo{year}{2019}), \bibinfo{pages}{429—439}.
\newblock
\showISSN{1554-351X}


\bibitem[Long et~al\mbox{.}(2018)]%
        {long2018conditional}
\bibfield{author}{\bibinfo{person}{Mingsheng Long}, \bibinfo{person}{Zhangjie
  Cao}, \bibinfo{person}{Jianmin Wang}, {and} \bibinfo{person}{Michael~I
  Jordan}.} \bibinfo{year}{2018}\natexlab{}.
\newblock \showarticletitle{Conditional adversarial domain adaptation}. In
  \bibinfo{booktitle}{\emph{Advances in Neural Information Processing
  Systems}}. \bibinfo{pages}{1645--1655}.
\newblock


\bibitem[Loshchilov and Hutter(2019)]%
        {loshchilov2018decoupled}
\bibfield{author}{\bibinfo{person}{Ilya Loshchilov} {and}
  \bibinfo{person}{Frank Hutter}.} \bibinfo{year}{2019}\natexlab{}.
\newblock \showarticletitle{Decoupled Weight Decay Regularization}. In
  \bibinfo{booktitle}{\emph{International Conference on Learning
  Representations}}.
\newblock
\urldef\tempurl%
\url{https://openreview.net/forum?id=Bkg6RiCqY7}
\showURL{%
\tempurl}


\bibitem[Mathur et~al\mbox{.}(2020)]%
        {mathur2020libriadapt}
\bibfield{author}{\bibinfo{person}{Akhil Mathur}, \bibinfo{person}{Fahim
  Kawsar}, \bibinfo{person}{Nadia Berthouze}, {and}
  \bibinfo{person}{Nicholas~D. Lane}.} \bibinfo{year}{2020}\natexlab{}.
\newblock \showarticletitle{Libri-Adapt: a New Speech Dataset for Unsupervised
  Domain Adaptation}. In \bibinfo{booktitle}{\emph{ICASSP 2020 - 2020 IEEE
  International Conference on Acoustics, Speech and Signal Processing
  (ICASSP)}}. \bibinfo{pages}{7439--7443}.
\newblock


\bibitem[P\'{e}rez-Rosas et~al\mbox{.}(2015)]%
        {perezrosas2015rlt}
\bibfield{author}{\bibinfo{person}{Ver\'{o}nica P\'{e}rez-Rosas},
  \bibinfo{person}{Mohamed Abouelenien}, \bibinfo{person}{Rada Mihalcea}, {and}
  \bibinfo{person}{Mihai Burzo}.} \bibinfo{year}{2015}\natexlab{}.
\newblock \showarticletitle{Deception Detection using Real-life Trial Data}. In
  \bibinfo{booktitle}{\emph{Proceedings of the 2015 ACM on International
  Conference on Multimodal Interaction}} (Seattle, Washington, USA)
  \emph{(\bibinfo{series}{ICMI '15})}. \bibinfo{publisher}{Association for
  Computing Machinery}, \bibinfo{address}{New York, NY, USA}.
\newblock
\showISBNx{9781450339124}


\bibitem[P{\'e}rez-Rosas et~al\mbox{.}(2015)]%
        {perez2015verbal}
\bibfield{author}{\bibinfo{person}{Ver{\'o}nica P{\'e}rez-Rosas},
  \bibinfo{person}{Mohamed Abouelenien}, \bibinfo{person}{Rada Mihalcea},
  \bibinfo{person}{Yao Xiao}, \bibinfo{person}{CJ Linton}, {and}
  \bibinfo{person}{Mihai Burzo}.} \bibinfo{year}{2015}\natexlab{}.
\newblock \showarticletitle{Verbal and Nonverbal Clues for Real-life Deception
  Detection}. In \bibinfo{booktitle}{\emph{Proceedings of the 2015 Conference
  on Empirical Methods in Natural Language Processing}},
  \bibfield{editor}{\bibinfo{person}{Llu{\'i}s M{\`a}rquez},
  \bibinfo{person}{Chris Callison-Burch}, {and} \bibinfo{person}{Jian Su}}
  (Eds.). \bibinfo{publisher}{Association for Computational Linguistics},
  \bibinfo{address}{Lisbon, Portugal}, \bibinfo{pages}{2336--2346}.
\newblock


\bibitem[P{\'e}rez-Rosas and Mihalcea(2015)]%
        {perez2015experiments}
\bibfield{author}{\bibinfo{person}{Ver{\'o}nica P{\'e}rez-Rosas} {and}
  \bibinfo{person}{Rada Mihalcea}.} \bibinfo{year}{2015}\natexlab{}.
\newblock \showarticletitle{Experiments in Open Domain Deception Detection}. In
  \bibinfo{booktitle}{\emph{Proceedings of the 2015 Conference on Empirical
  Methods in Natural Language Processing}},
  \bibfield{editor}{\bibinfo{person}{Llu{\'i}s M{\`a}rquez},
  \bibinfo{person}{Chris Callison-Burch}, {and} \bibinfo{person}{Jian Su}}
  (Eds.). \bibinfo{publisher}{Association for Computational Linguistics},
  \bibinfo{address}{Lisbon, Portugal}, \bibinfo{pages}{1120--1125}.
\newblock


\bibitem[Shankar et~al\mbox{.}(2018)]%
        {shankar2018generalizing}
\bibfield{author}{\bibinfo{person}{Shiv Shankar}, \bibinfo{person}{Vihari
  Piratla}, \bibinfo{person}{Soumen Chakrabarti}, \bibinfo{person}{Siddhartha
  Chaudhuri}, \bibinfo{person}{Preethi Jyothi}, {and} \bibinfo{person}{Sunita
  Sarawagi}.} \bibinfo{year}{2018}\natexlab{}.
\newblock \showarticletitle{Generalizing Across Domains via Cross-Gradient
  Training}. In \bibinfo{booktitle}{\emph{International Conference on Learning
  Representations}}.
\newblock


\bibitem[Shi et~al\mbox{.}(2020)]%
        {shi2020towards}
\bibfield{author}{\bibinfo{person}{Yichun Shi}, \bibinfo{person}{Xiang Yu},
  \bibinfo{person}{Kihyuk Sohn}, \bibinfo{person}{Manmohan Chandraker}, {and}
  \bibinfo{person}{Anil~K. Jain}.} \bibinfo{year}{2020}\natexlab{}.
\newblock \showarticletitle{Towards Universal Representation Learning for Deep
  Face Recognition}. In \bibinfo{booktitle}{\emph{Proceedings of the IEEE/CVF
  Conference on Computer Vision and Pattern Recognition (CVPR)}}.
\newblock


\bibitem[Soldner et~al\mbox{.}(2019)]%
        {soldner2019boxoflies}
\bibfield{author}{\bibinfo{person}{Felix Soldner},
  \bibinfo{person}{Ver{\'o}nica P{\'e}rez-Rosas}, {and} \bibinfo{person}{Rada
  Mihalcea}.} \bibinfo{year}{2019}\natexlab{}.
\newblock \showarticletitle{Box of Lies: Multimodal Deception Detection in
  Dialogues}. In \bibinfo{booktitle}{\emph{Proceedings of the 2019 Conference
  of the North {A}merican Chapter of the Association for Computational
  Linguistics: Human Language Technologies, Volume 1 (Long and Short Papers)}},
  \bibfield{editor}{\bibinfo{person}{Jill Burstein}, \bibinfo{person}{Christy
  Doran}, {and} \bibinfo{person}{Thamar Solorio}} (Eds.).
  \bibinfo{publisher}{Association for Computational Linguistics},
  \bibinfo{address}{Minneapolis, Minnesota}, \bibinfo{pages}{1768--1777}.
\newblock


\bibitem[Sun and Saenko(2016)]%
        {sun2016coral}
\bibfield{author}{\bibinfo{person}{Baochen Sun} {and} \bibinfo{person}{Kate
  Saenko}.} \bibinfo{year}{2016}\natexlab{}.
\newblock \showarticletitle{Deep CORAL: Correlation Alignment for Deep Domain
  Adaptation}. In \bibinfo{booktitle}{\emph{Computer Vision -- ECCV 2016
  Workshops}}, \bibfield{editor}{\bibinfo{person}{Gang Hua} {and}
  \bibinfo{person}{Herv{\'e} J{\'e}gou}} (Eds.). \bibinfo{publisher}{Springer
  International Publishing}, \bibinfo{address}{Cham},
  \bibinfo{pages}{443--450}.
\newblock
\showISBNx{978-3-319-49409-8}


\bibitem[Toisoul et~al\mbox{.}(2021)]%
        {toisoul2021estimation}
\bibfield{author}{\bibinfo{person}{Antoine Toisoul}, \bibinfo{person}{Jean
  Kossaifi}, \bibinfo{person}{Adrian Bulat}, \bibinfo{person}{Georgios
  Tzimiropoulos}, {and} \bibinfo{person}{Maja Pantic}.}
  \bibinfo{year}{2021}\natexlab{}.
\newblock \showarticletitle{Estimation of continuous valence and arousal levels
  from faces in naturalistic conditions}.
\newblock \bibinfo{journal}{\emph{Nature Machine Intelligence}}
  (\bibinfo{year}{2021}).
\newblock


\bibitem[Tsai et~al\mbox{.}(2019)]%
        {tsai2019multimodal}
\bibfield{author}{\bibinfo{person}{Yao-Hung~Hubert Tsai},
  \bibinfo{person}{Shaojie Bai}, \bibinfo{person}{Paul~Pu Liang},
  \bibinfo{person}{J.~Zico Kolter}, \bibinfo{person}{Louis-Philippe Morency},
  {and} \bibinfo{person}{Ruslan Salakhutdinov}.}
  \bibinfo{year}{2019}\natexlab{}.
\newblock \showarticletitle{Multimodal Transformer for Unaligned Multimodal
  Language Sequences}. In \bibinfo{booktitle}{\emph{Proceedings of the 57th
  Annual Meeting of the Association for Computational Linguistics}},
  \bibfield{editor}{\bibinfo{person}{Anna Korhonen}, \bibinfo{person}{David
  Traum}, {and} \bibinfo{person}{Llu{\'i}s M{\`a}rquez}} (Eds.).
  \bibinfo{publisher}{Association for Computational Linguistics},
  \bibinfo{address}{Florence, Italy}.
\newblock


\bibitem[Vance et~al\mbox{.}(2022)]%
        {vance2022deception}
\bibfield{author}{\bibinfo{person}{Nathan Vance}, \bibinfo{person}{Jeremy
  Speth}, \bibinfo{person}{Siamul Khan}, \bibinfo{person}{Adam Czajka},
  \bibinfo{person}{Kevin~W. Bowyer}, \bibinfo{person}{Diane Wright}, {and}
  \bibinfo{person}{Patrick Flynn}.} \bibinfo{year}{2022}\natexlab{}.
\newblock \showarticletitle{Deception Detection and Remote Physiological
  Monitoring: A Dataset and Baseline Experimental Results}.
\newblock \bibinfo{journal}{\emph{IEEE Transactions on Biometrics, Behavior,
  and Identity Science}} \bibinfo{volume}{4}, \bibinfo{number}{4}
  (\bibinfo{year}{2022}), \bibinfo{pages}{522--532}.
\newblock


\bibitem[Xu et~al\mbox{.}(2020)]%
        {xu2020adversarial}
\bibfield{author}{\bibinfo{person}{Minghao Xu}, \bibinfo{person}{Jian Zhang},
  \bibinfo{person}{Bingbing Ni}, \bibinfo{person}{Teng Li},
  \bibinfo{person}{Chengjie Wang}, \bibinfo{person}{Qi Tian}, {and}
  \bibinfo{person}{Wenjun Zhang}.} \bibinfo{year}{2020}\natexlab{}.
\newblock \showarticletitle{Adversarial Domain Adaptation with Domain Mixup}.
  In \bibinfo{booktitle}{\emph{The Thirty-Fourth AAAI Conference on Artificial
  Intelligence}}. \bibinfo{publisher}{AAAI Press}, \bibinfo{pages}{6502--6509}.
\newblock


\bibitem[Yang et~al\mbox{.}(2021)]%
        {yang2021constructing}
\bibfield{author}{\bibinfo{person}{Jun-Teng Yang}, \bibinfo{person}{Guei-Ming
  Liu}, {and} \bibinfo{person}{Scott C-H Huang}.}
  \bibinfo{year}{2021}\natexlab{}.
\newblock \showarticletitle{Constructing robust emotional state-based feature
  with a novel voting scheme for multi-modal deception detection in videos}.
\newblock \bibinfo{journal}{\emph{arXiv preprint arXiv:2104.08373}}
  (\bibinfo{year}{2021}).
\newblock


\bibitem[Zhang et~al\mbox{.}(2015)]%
        {zhang2015multisource}
\bibfield{author}{\bibinfo{person}{Kun Zhang}, \bibinfo{person}{Mingming Gong},
  {and} \bibinfo{person}{Bernhard Schoelkopf}.}
  \bibinfo{year}{2015}\natexlab{}.
\newblock \showarticletitle{Multi-Source Domain Adaptation: A Causal View}.
\newblock \bibinfo{journal}{\emph{Proceedings of the AAAI Conference on
  Artificial Intelligence}} \bibinfo{volume}{29}, \bibinfo{number}{1}
  (\bibinfo{date}{Feb.} \bibinfo{year}{2015}).
\newblock


\bibitem[Zhao et~al\mbox{.}(2024)]%
        {zhao2024more}
\bibfield{author}{\bibinfo{person}{Sicheng Zhao}, \bibinfo{person}{Hui Chen},
  \bibinfo{person}{Hu Huang}, \bibinfo{person}{Pengfei Xu}, {and}
  \bibinfo{person}{Guiguang Ding}.} \bibinfo{year}{2024}\natexlab{}.
\newblock \showarticletitle{More is better: deep domain adaptation with
  multiple sources}. In \bibinfo{booktitle}{\emph{Proceedings of the
  Thirty-Third International Joint Conference on Artificial Intelligence}}
  (Jeju, Korea) \emph{(\bibinfo{series}{IJCAI '24})}. Article
  \bibinfo{articleno}{923}, \bibinfo{numpages}{9}~pages.
\newblock
\showISBNx{978-1-956792-04-1}


\bibitem[Zhao et~al\mbox{.}(2025)]%
        {zhao2025mmda}
\bibfield{author}{\bibinfo{person}{Sicheng Zhao}, \bibinfo{person}{Jing Jiang},
  \bibinfo{person}{Wenbo Tang}, \bibinfo{person}{Jiankun Zhu},
  \bibinfo{person}{Hui Chen}, \bibinfo{person}{Pengfei Xu},
  \bibinfo{person}{Björn~W. Schuller}, \bibinfo{person}{Jianhua Tao},
  \bibinfo{person}{Hongxun Yao}, {and} \bibinfo{person}{Guiguang Ding}.}
  \bibinfo{year}{2025}\natexlab{}.
\newblock \showarticletitle{Multi-source multi-modal domain adaptation}.
\newblock \bibinfo{journal}{\emph{Information Fusion}}  \bibinfo{volume}{117}
  (\bibinfo{year}{2025}), \bibinfo{pages}{102862}.
\newblock
\showISSN{1566-2535}


\bibitem[Zhao et~al\mbox{.}(2020)]%
        {zhao2020multi}
\bibfield{author}{\bibinfo{person}{Sicheng Zhao}, \bibinfo{person}{Bo Li},
  \bibinfo{person}{Pengfei Xu}, {and} \bibinfo{person}{Kurt Keutzer}.}
  \bibinfo{year}{2020}\natexlab{}.
\newblock \showarticletitle{Multi-source domain adaptation in the deep learning
  era: A systematic survey}.
\newblock \bibinfo{journal}{\emph{arXiv preprint arXiv:2002.12169}}
  (\bibinfo{year}{2020}).
\newblock


\bibitem[Şen et~al\mbox{.}(2022)]%
        {sen2022multimodal}
\bibfield{author}{\bibinfo{person}{M.~Umut Şen}, \bibinfo{person}{Verónica
  Pérez-Rosas}, \bibinfo{person}{Berrin Yanikoglu}, \bibinfo{person}{Mohamed
  Abouelenien}, \bibinfo{person}{Mihai Burzo}, {and} \bibinfo{person}{Rada
  Mihalcea}.} \bibinfo{year}{2022}\natexlab{}.
\newblock \showarticletitle{Multimodal Deception Detection Using Real-Life
  Trial Data}.
\newblock \bibinfo{journal}{\emph{IEEE Transactions on Affective Computing}}
  \bibinfo{volume}{13}, \bibinfo{number}{1} (\bibinfo{year}{2022}),
  \bibinfo{pages}{306--319}.
\newblock


\end{thebibliography}

\appendix

\section{Implementation Details}
\label{sec:appendix_implementation}
All experiments are performed on 1 H800 GPUs with CUDA 12.4 and Pytorch . We utilize AdamW \cite{loshchilov2018decoupled} as the training optimizer with learning rate as $1e-4$ and weight decay as $5e-5$. The batch size is set as $32$ and the maximum epoch is set as $20$. The layer number of multimodal fusion model MulT is kept as $3$ and the dimension for unimodal and multimodal representations are set as $512$ and $128$. The computational FLOPs (Floating Point Operations per second) of the proposed model are $60.25$ GFLOPs and the total model parameters are $33.59$ M.

\section{More Dataset Distribution Details}
\label{sec:appendix_dataset}
\begin{figure*}[htbp]
    \centering 
    \subfloat[Mean on Expression classes] {
    \includegraphics[width=0.62\columnwidth]{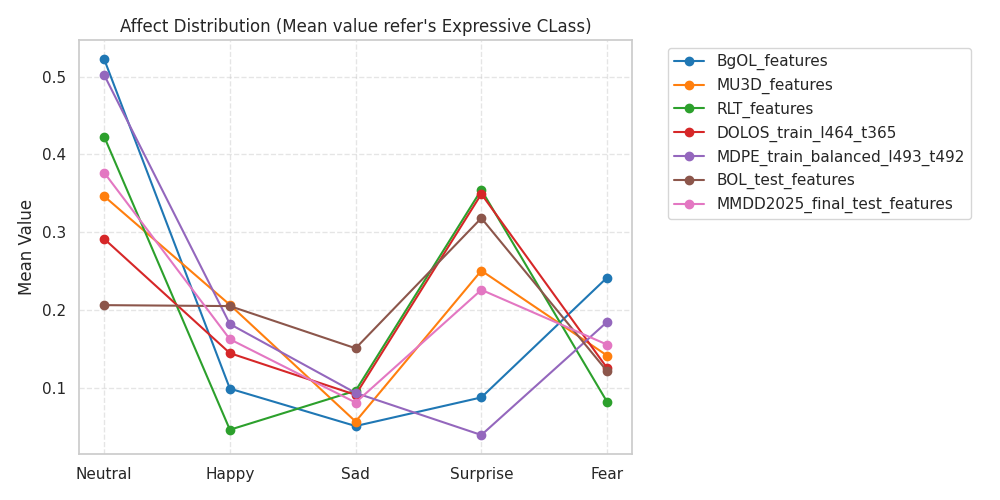}
    } 
    \subfloat[Valence Value $F_f$] {
    \includegraphics[width=0.72\columnwidth]{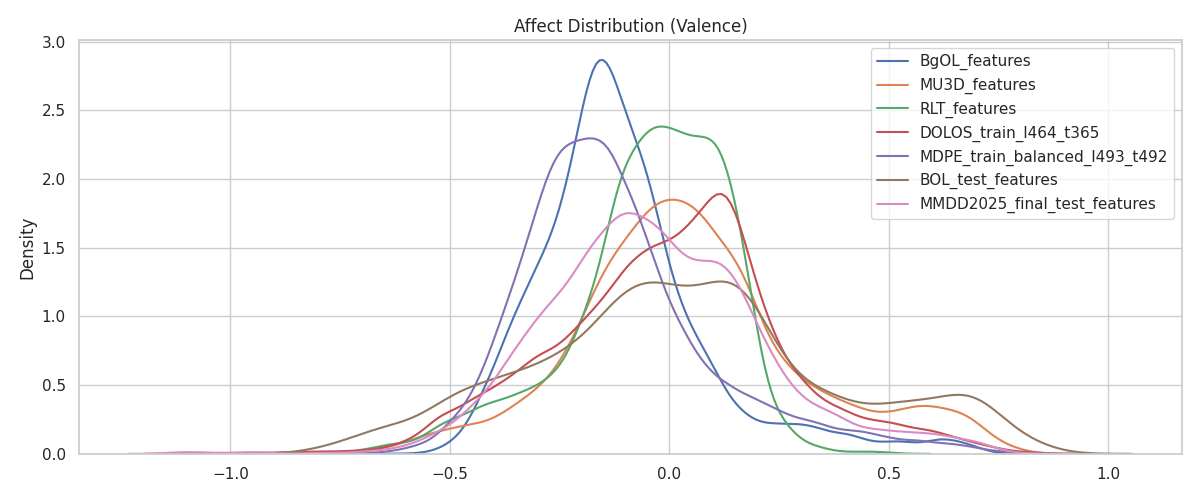}     
    }
    \subfloat[Arousal Value $F_b$] {
    \includegraphics[width=0.72\columnwidth]{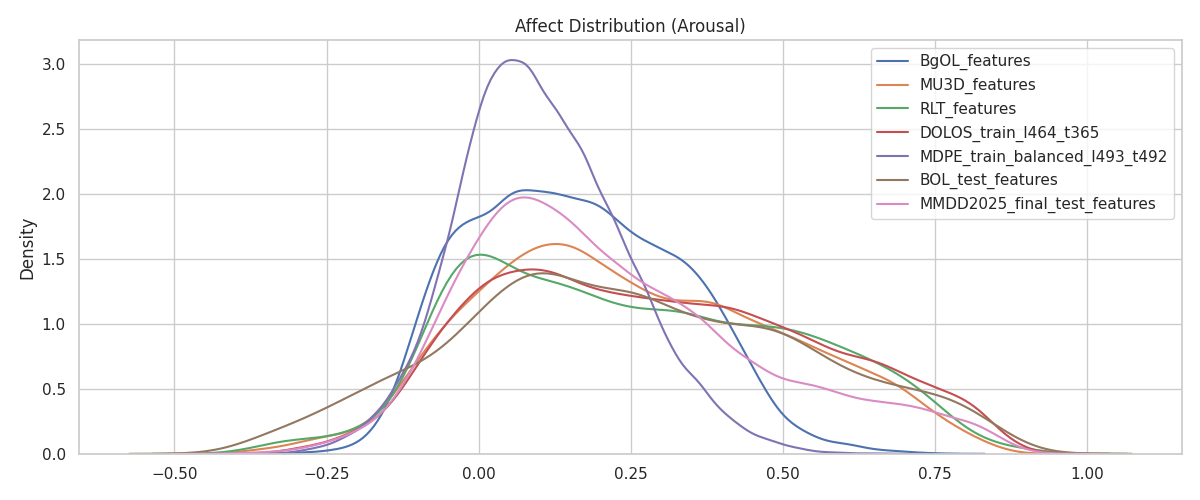}
    }
    \caption{Domain shift visualization for affect features in behavior features in different dataset domains.}
    \label{vis_affect}
\end{figure*}

\begin{figure}[htbp]
	\centering 
	\includegraphics[scale=0.18] {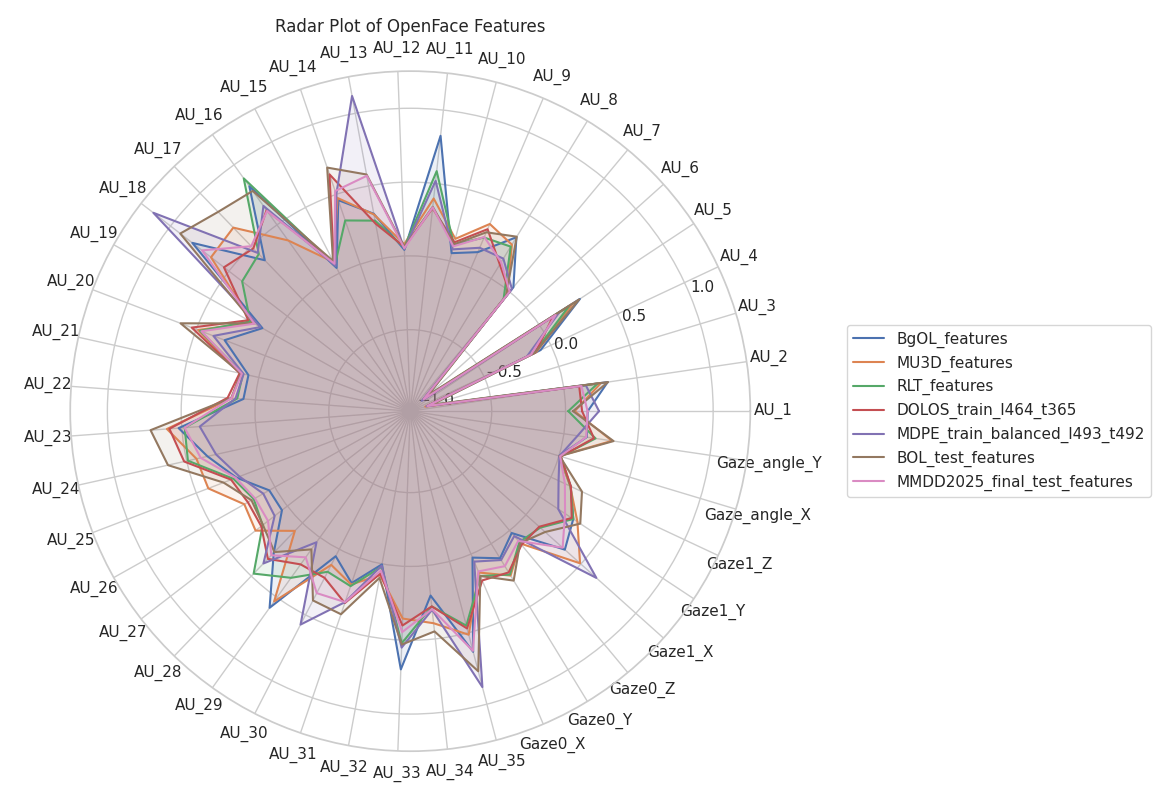}
	\caption{Domain shift visualization for gaze and AUs features in behavior features in different dataset domains. }
	\label{vis_openface}
\end{figure}

We visualize the distribution of behavior features with t-SNE algorithm and statistical analysis to further validate the significance of cross-domain shift as shown in Figure \ref{vis_domain_shift}, \ref{vis_affect}, and \ref{vis_openface}. 

For audio spectrogram features, dataset domains form almost completely separate clusters, showing large modality-specific domain gaps. Although the visual face features present some mixing, but there are still visible clusters corresponding to certain domains, indicating partial but incomplete alignment. 

For behavior features show more overlap in one single cluster, but small separations persist in the latent space. Thus, we further visualize each dimension of the behavior features across diverse datasets in Figure \ref{vis_affect}, since there are corresponding semantics information extracted by Emonet (7-dimensional Affect features, with 5 dimension expression classes, 1 dimension arousal value, 1 dimension valence value) \cite{toisoul2021estimation} and Openface (43-dimensional with 8 dimension gaze and 35 dimension AUs value) \cite{amos2016openface}.  The mean values differ substantially across dimension and the expression class imbalance vary greatly across domains. The valence and arousal distributions also vary notably, BgOL and MDPE datasets show sharper peaks and distinctive ranges, while other datasets have flatter distributions. Besides, the radar plot visualized in Figure \ref{vis_openface} highlights variations in facial behavior distributions including AU intensity and gaze direction. These discrepancies indicate that emotion intensity and class balance are not consistent across domains, making cross-domain adaption harder. 

Overall, these results reveal that both low-level acoustic-visual features and high-level affective features suffer from domain shift, indicating the effectiveness of MMPDA for multimodal domain adaptation.

\section{Limitation}
\label{sec:appendix_limitation}
While MMPDA effectively reduces modality gaps and domain shifts at the multimodal level in a progressive manner, it does not explicitly adapt multi-scale unimodal features across different domains. Both unimodal and multimodal extraction process may undergo distinct types of domain shift. Without aligning these unimodal features before fusion, residual mismatches could weaken the final multimodal representation and limit generalization.

In addition, the joint training of unimodal and multimodal tasks may results in optimization conflicts, where adapting one modality too aggressively may harm the synergy between modalities. Balancing these objectives may require adaptive gradient regularization or weighting strategies to ensure their respective optimization process. Finally, the current evaluation only covers selected source–target setups, leaving open challenges of how MMPDA performs when each dataset is used as the target domain with various multi-source combinations. More exhaustive experiments on diverse combinations of source and target domains could reveal transferability ability and cross-domain influence.


\end{document}